\documentclass[11pt,a4paper]{article}
\usepackage[hyperref]{emnlp2020}
\usepackage{times}
\usepackage{latexsym}

\usepackage{microtype}

\usepackage{soul}
\usepackage{xcolor}
\usepackage{graphicx}
\usepackage{booktabs}
\usepackage{multirow}
\usepackage{placeins}
\usepackage{tablefootnote}
\usepackage{threeparttable}
\usepackage{diagbox}
\usepackage{dialogue}
\usepackage[normalem]{ulem}
\usepackage{mwe}
\usepackage{array}
\usepackage{etoolbox}
\usepackage{url}
\usepackage{calc}
\usepackage{tcolorbox}
\usepackage{setspace}
\usepackage{amsmath}
\usepackage{amsfonts}
\usepackage{mathtools}
\usepackage{leftidx}
\usepackage{ulem}
\usepackage{slashbox}
\usepackage{textcomp}
\usepackage[caption = false]{subfig}
\usepackage[colorinlistoftodos,prependcaption,textsize=tiny]{todonotes}
\newif\ifsubmissionready
\submissionreadyfalse
\def\readytosubmit{\global\submissionreadytrue}

\aclfinalcopy % Uncomment this line for the final submission
\readytosubmit % Uncomment this to hide comments and annotations before submitting

\newlength{\DepthReference}
\newlength{\HeightReference}
\newlength{\Width}%

\newcommand{\MyColorBox}[2][red]%
{%
    \settowidth{\Width}{#2}%
    \colorbox{#1}%
    {%      
        \raisebox{-\DepthReference}%
        {%
                \parbox[b][\HeightReference+\DepthReference][c]{\Width}{\centering#2}%
        }%
    }%
}

\definecolor{blue}{RGB}{0, 93, 170}			%Go Big Blue!
\definecolor{darkgreen}{RGB}{0, 102, 0}
\definecolor{lightskyblue}{rgb}{0.53, 0.81, 0.98}
\definecolor{lightgreen}{rgb}{0.56, 0.93, 0.56}
\definecolor{palegoldenrod}{rgb}{0.93, 0.91, 0.67}
\newcommand{\hmed}[1]{\MyColorBox[red]{#1}}
\newcommand{\hstartstop}[1]{\MyColorBox[lightskyblue]{#1}}

\newcommand{\hroute}[1]{\MyColorBox[lightgreen]{#1}}

\newcommand{\hfreq}[1]{\MyColorBox[palegoldenrod]{#1}}

\newcommand{\newlyadded}[1]{\ifsubmissionready#1\else\textcolor{darkgreen}{#1}\fi}

\newcommand{\hide}[1]{}
\let\oldsout\sout
\renewcommand{\sout}[1]{\ifsubmissionready\else\oldsout{#1}\fi}
\newcommand{\freq}{\emph{frequency}}
\newcommand{\route}{\emph{route}}
\newcommand{\startstop}{\emph{change}}

\DeclareMathOperator*{\argmax}{arg\,max}
\DeclareMathOperator*{\argmin}{arg\,min}

\title{Weakly Supervised Medication Regimen Extraction \\ from Medical Conversations}

\author{Dhruvesh Patel\thanks{~~Work done as an intern at Abridge AI Inc.}\\
  College of Information and Computer Sciences \\
  University of Massachusetts Amherst \\
  \texttt{dhruveshpate@cs.umass.edu} 
  \AND
  Sandeep Konam\\
  Abridge AI Inc.\\
  \texttt{san@abridge.com} \\\And
  Sai P. Selvaraj \\
  Abridge AI Inc.\\
  \texttt{prabhakarsai@abridge.com }
  }

\date{}

\begin{document}
\maketitle

\begin{abstract}
\end{abstract}
Automated Medication Regimen (MR) extraction from medical conversations can not only improve recall and help patients follow through with their care plan, but also reduce the documentation burden for doctors. In this paper, we focus on extracting spans for
\freq, \route~and \startstop, corresponding to medications discussed in the conversation. We first describe a unique dataset of annotated doctor-patient conversations and then
present a weakly supervised model architecture that can perform span extraction using noisy classification data.
The model utilizes an attention bottleneck inside a classification model to perform the extraction.
We experiment with several variants of attention scoring and projection functions and propose a novel transformer-based attention scoring function (TAScore). The proposed combination of TAScore and Fusedmax projection achieves a 10 point increase in Longest Common Substring F1 compared to the baseline of additive scoring plus softmax projection.

\makeatletter
\def\blfootnote{\gdef\@thefnmark{}\@footnotetext}
\makeatother

\section{Introduction}
\label{sec:intro}
\blfootnote{Proceedings of the Clinical Natural Language Processing Workshop, EMNLP, 2020.}

Patients forget 40-80\% of the medical information provided by healthcare practitioners immediately \citep{doi:10.1080/03610739608254020} and misconstrue 48\% of what they think they remembered \citep{10.1093/rheumatology/18.1.18}, and this adversely affects patient adherence. Automatically extracting information from doctor-patient conversations can help patients correctly recall doctor’s instructions and improve compliance with the care plan \cite{tsulukidze2014providing}. On the other hand, clinicians spend up to 49.2\% of their overall time on EHR and desk work, and only 27.0\% of their total time on direct clinical face time with patients \citep{sinsky2016allocation}. Increased data management work is also correlated with increased doctor burnout \citep{kumar2016burnout}.  
Information extracted from medical conversations can also aid doctors in their documentation work \citep{rajkomar2019automatically,ben2020soap}, allow them to spend more face time with the patients, and build better relationships.

\begin{figure}[!t]
\begin{tcolorbox}[width=\columnwidth]
\begin{singlespacing}
{\setstretch{1.2}
    \begin{dialogue}
    \small
        \speak{DR}  Limiting your alcohol consumption is important, so, and, um, so, you know, I would recommend \hmed{vitamin D\textsuperscript{1}} \hstartstop{to be taken\textsuperscript{1}}. Have you had \hmed{Fosamax\textsuperscript{2}} before?
        \speak{PT}  I think my mum did.
        \speak{DR} Okay, \hmed{Fosamax\textsuperscript{2}}, \hstartstop{you take\textsuperscript{2}} \hfreq{one \hroute{pill\textsuperscript{2}} on Monday and one on Thursday\textsuperscript{2}}.
        \speak{DR} Do you use much caffine?
        \speak{PT} No, none.
        \speak{DR} Okay, \hstartstop{this is\textsuperscript{3}} \hmed{Actonel\textsuperscript{3}} and it's \hfreq{one \hroute{tablet\textsuperscript{3}} once a month\textsuperscript{3}}.
        \speak{DR} Do you get a one month or a three months supply in your prescriptions?
    \end{dialogue}
}
\end{singlespacing}
\end{tcolorbox}
\caption{An example excerpt from a doctor-patient conversation transcript. Here, there are three \hmed{medications} mentioned indicated by the superscript. The extracted attributes, \hstartstop{change}, \hroute{route} and \hfreq{frequency}, for each medications are also shown. }
\label{fig:dialogue}
\end{figure}

In this work, we focus on extracting Medication Regimen (MR) information \cite{du2019learning, selvaraj2019medication} from the doctor-patient conversations. Specifically, we extract three attributes, i.e., \freq, \route~and \startstop, corresponding to medications discussed in the conversation (Figure \ref{fig:dialogue}). Medication Regimen information can help doctors with medication orders cum renewals, medication reconciliation, verification of reconciliations for errors, and other medication-centered EHR documentation tasks. It can also improve patient engagement, transparency and better compliance with the care plan \cite{tsulukidze2014providing, grande2017digital}.

% Classification data is easy to obtain 
MR attribute information present in a conversation can be obtained as spans in text (Figure \ref{fig:dialogue}) or can be categorized into classification labels (Table \ref{tab:dataset example}). While the classification labels are easy to obtain at scale in an automated manner -- for instance, by pairing conversations with billing codes or medication orders -- 
they can be noisy and can result in a prohibitively large number of classes. 
Classification labels go through normalization and disambiguation, often resulting in label names which are very different from the phrases used in the conversation. This process leads to a loss of granular information present in the text (see, for example, row 2 in Table \ref{tab:dataset example}).
Span extraction, on the other hand, alleviates this issue as the outputs are actual spans in the conversation. However, span extraction annotations are relatively hard to come by and are time-consuming to annotate manually.
Hence, in this work, we look at the task of MR attribute span extraction from doctor-patient conversation using weak supervision provided by the noisy classification labels. 

% HOW
The main contributions of this work are as follows.
We present a way of setting up an MR attribute extraction task from noisy classification data (Section \ref{sec:task and data}). We propose a weakly supervised model architecture which utilizes attention bottleneck inside a classification model to perform span extraction (Section \ref{sec:approach} \& \ref{sec:model}). In order to favor sparse and contiguous extractions, we experiment with two variants of attention projection functions (Section \ref{sec:projection function}), namely, softmax and Fusedmax \cite{fusedmax}.
Further, we propose a novel transformer-based attention scoring function TAScore (Section \ref{sec:scoring function}). The combination of TAScore and Fusedmax achieves significant improvements in extraction performance over a phrase-based (22 LCSF1 points) and additive softmax attention (10 LCSF1 points) baselines.

\section{Medication Regimen (MR) using Weak Supervision}
\label{sec:task and data}

Medication Regimen (MR) consists of information about a prescribed medication akin to attributes of an entity. In this work, we specifically focus on \freq, \route~of the medication and any \startstop~in the medication's \emph{dosage} or \freq~ as shown in Figure \ref{fig:dialogue}.  
 For example, given the conversation excerpt and the medication ``Fosamax'' as shown in Figure \ref{fig:dialogue}, the model needs to extract the spans ``one pill on Monday and one on Thursday'', ``pill'' and ``you take'' for attributes \freq, \route ~and \startstop, respectively. The major challenge, however, is to perform the attribute span extraction using noisy classification labels with very few or no span-level labels. The rest of this section describes the dataset used for this task.

\begin{table}[]
\centering
\resizebox{\columnwidth}{!}{%
{\small
\begin{tabular}{|l|l|} 
\toprule
Attribute                 & Normalized Classes                                                                                                                                                                                                                                                                                                   \\ 
\midrule
\freq      & \begin{tabular}[c]{@{}l@{}}Daily \textbar{} Every morning \textbar{} At Bedtime \textbar{}\\Twice a day \textbar{} Three times a day \textbar{} Every six hours \textbar{}\\Every week \textbar{} Twice a week \textbar{} Three times a week \textbar{}\\Every month \textbar{} Other \textbar{} None \end{tabular}  \\ 
\midrule
\route     & \begin{tabular}[c]{@{}l@{}}Pill \textbar{} Injection \textbar{} Topical cream \textbar{} Nasal spray \textbar{}\\Medicated patch \textbar{} Ophthalmic solution \textbar{} Inhaler \textbar{}\\Oral solution \textbar{} Other \textbar{} None \end{tabular}                                                          \\ 
\midrule
\startstop & Take \textbar{} Stop \textbar{} Increase \textbar{} Decrease \textbar{} None \textbar{} Other                                                                                                                                                                                                                        \\
\bottomrule
\end{tabular}
}
}
\caption{The normalized labels in the classification data.}
\label{tab:class labels}
\end{table}

\begin{table*}[]
\centering
\resizebox{\textwidth}{!}{%
\begin{tabular}{@{}lcccc@{}}
\toprule
 \multirow{2}{*}{\emph{text}} & \multirow{2}{*}{\emph{medication}} &\multicolumn{3}{c}{Classification labels}                      \\ \cmidrule(l){3-5} 
                      &               & \multicolumn{1}{l}{\freq} & \route & \startstop \\ \midrule

  \begin{tabular}[c]{@{}l@{}}{\small$\dots$ I would recommend vitamin D to be taken.}\\[-2mm] {\small Have you had Fosamax before?$\dots$}\end{tabular} & { vitamin D} &
  { none} &
  \multicolumn{1}{c}{{ none} } &
  take \\\midrule

  \begin{tabular}[c]{@{}l@{}}{\small$\dots$ I think my mum did. Okay, Fosamax, you take one pill on Monday}\\[-2mm] {\small and one on Thursday. Do you have much caffine? No, none$\dots$}\end{tabular}  & Fosamax &
  Twice a week &
  \multicolumn{1}{c}{pill} &
  take \\\midrule

  \begin{tabular}[c]{@{}l@{}}{\small Do you have much caffine? No, none. Okay, this is Actonel and it's,}\\[-2mm] {\small one tablet once a month.$\dots$}
  \end{tabular} & Actonel &
  Once a month &
  \multicolumn{1}{c}{pill} &
  take \\
  \bottomrule
\end{tabular}
}
\caption{Classification examples resulting from the conversation shown in Figure \ref{fig:dialogue}.}
\label{tab:dataset example}
\end{table*}

\subsection{Data}

The data used in this paper comes from a collection of human transcriptions of 63000 fully-consented and de-identified doctor-patient conversations. 
A total of 57000 conversations were randomly selected to construct the training (and dev) conversation pool and the remaining 6000 conversations were reserved as the test pool.

\smallskip
\noindent\textbf{The classification dataset: }All the conversations are annotated with MR tags by expert human annotators. Each set of MR tags consists of the \emph{medication} name and its corresponding attributes \freq, \route~ and \startstop, which are normalized free-form instructions in natural language phrases corresponding to each of the three attributes (see Table \ref{tab:phrase based phrases} in \ref{app:phrase based}). 
Each set of MR tags is grounded to a contiguous window of utterances' \emph{text},\footnote{The \emph{text} includes both the spoken words and the speaker information.} around a medication mention as evidence for that set. 
Hence, each set of grounded MR tags can be written as $<$\emph{medication}, \emph{text}, \freq, \route, \startstop$>$, where the last three entries correspond to the three MR attributes. 

The free-form instructions for each attribute in the MR tags are normalized and categorized into manageable number of classification labels to avoid long tail and overlapping classes. 
This process results in classes shown in Table \ref{tab:class labels}.\footnote{The detailed explanation for each of the classes can be found in Table \ref{tab:class labels detailed} in Appendix \ref{app:data}.}
As an illustration, this annotation process when applied to the conversation piece shown in Figure \ref{fig:dialogue} would result in the three data points shown in Table \ref{tab:dataset example}. 
Using this procedure on both the training and test conversation pools, we obtain 45,059 training, 11,212 validation and 5,458 test classification data points.\footnote{The dataset statistics are given in Appendix \ref{app:data}.}

\smallskip
\noindent\textbf{The extraction dataset: }
Since the goal is to extract spans related to MR attributes, we would ideally need a dataset with span annotations to perform this task in a fully supervised manner. However, span annotation is laborious and expensive. Hence, we re-purpose the classification dataset (along with its classification labels) to perform the task of span extraction using weak supervision. We also manually annotate a small fraction of the train, validation and test sets (150, 150 and 500 data-points respectively) for attribute spans to see the effect of supplying a small number of strongly supervised instances on the performance of the model. 
In order to have a good representation of all the classes in the test set, we increase the sampling weight of data-points which have rare classes. Hence, our test set is relatively more difficult compared to a random sample of 500 data-points.
All the results are reported on our test set of 500 difficult data-points annotated for attribute spans.

For annotating attribute spans, the annotators were given instructions to mark spans which provide minimally sufficient and natural evidence for the already annotated attribute class as described below.

\smallskip
\noindent\textbf{Sufficiency:} 
Given only the annotated span for a particular attribute, one should be able to predict the correct classification label. This aims to encourage the attribute spans to cover all distinguishing information for that attribute.

\noindent\textbf{Minimality: }Peripheral words which can be replaced with other words without changing the attribute's classification label should not be included in the extracted span. This aims to discourage marking entire utterances as attribute spans.

\noindent\textbf{Naturalness: }The marked span(s) if presented to a human should sound like complete English phrases (if it has multiple tokens) or a meaningful word if it has only a single token. In essence, this means that the extractions should not drop stop words from within phrases. \newlyadded{This requirement aims to reduce the cognitive load on the human who uses the model's extraction output.}

\subsection{Challenges}
\label{sec:challenges}
Using medical conversations for information extraction is more challenging compared to written doctor notes because the spontaneity of conversation gives rise to a variety of speech patterns with disfluencies and interruptions. Moreover, the vocabulary can range from colloquial to medical jargon. 

In addition, we also have noise in our classification dataset with its main source being annotators' use of information outside the grounded \emph{text} window to produce the free-form tags. This happens in two ways. First, when the free-form MR instructions are written using evidence that was discussed elsewhere in the  conversation but is not present in the grounded \emph{text} window. Second, when the annotator uses their domain knowledge instead of using just the information in the grounded \emph{text} window -- for instance, when the \emph{route} of a \emph{medication} is not explicitly mentioned, the annotator might use the \emph{medication}`s common \emph{route} in their free-form instructions. 
Using manual analysis of the 800 data-points across the train, dev and test sets, we find that 22\% of \freq, 36\% of \route~and 15\% of \startstop~ classification labels, have this noise. 

In this work, our approach to extraction depends on the size of the auxiliary task's (classification) dataset to overcome above mentioned challenges.

\section{Background}
\label{sec:approach}
%While there is an on-going debate whether neural attention provides plausible and faithful explanations for model decisions \cite{attention-is-not-explanation, attention-is-not-not-explanation}, 
There have been several successful attempts to use neural attention \cite{Bahdanau2015NeuralMT} to extract information from text in an unsupervised manner \cite{neural-attn-for-aspect-extraction, neural-relation-extraction, yu2019beyond}.  
Attention scores provide a good proxy for importance of a particular token in a model. 
However, when there are multiple layers of attention, or if the encoder is too complex and trainable, the model no longer provides a way to produce 
%\emph{faithful} -- \sai{define faith}, importance 
reliable and faithful importance 
scores \cite{attention-is-not-explanation}. 

We argue that, in order to bring in the faithfulness, we need to create an attention bottleneck in our classification + extraction model. The attention bottleneck is achieved by employing an attention function which generates a set of attention weights over the encoded input tokens. Attention bottleneck forces the classifier \emph{to only see} the portions of input that pass through it, thereby enabling us to trade the classification performance for extraction performance and getting span extraction with weak supervision from classification labels.

In the rest of this section, we provide general background on neural attention and present its variants employed in this work. This is followed by the presentation of our complete model architecture in the subsequent sections.

\subsection{Neural Attention}
\label{sec:attention}
Given a query $\mathbf{q} \in \mathbb{R}^m$ and keys $\mathbf{K} \in \mathbb{R}^{l\times n}$, the attention function $\alpha \colon \mathbb{R}^m \times   \mathbb{R}^{l\times n} \to \Delta^l$ is composed of two functions: a scoring function $\mathcal{S} \colon \mathbb{R}^m \times   \mathbb{R}^{l\times n} \to \mathbb{R}^l$ which produces unnormalized importance scores, and a projection function $\Pi \colon \mathbb{R}^l \to \Delta^l$ which normalizes these scores by projecting them to an $(l-1)$-dimensional probability simplex.\footnote{Throughout this work $l$ represents the sequence length dimension and $\Delta^l=\{\mathbf{x} \in \mathbb{R}^l ~|~ \mathbf{x}>0, \Vert \mathbf{x}\Vert_1 =1 \}$ represents a probability simplex.}

%We experiment with two scoring functions: multi-layer transformer \cite{transformers} and additive attention \cite{Bahdanau2015NeuralMT}. We use fusedmax \cite{fusedmax}, a regularized sparse version of softmax as the normalization function. \TODO{Add the projection equation. Or some other details about fusedmax.}

\subsubsection{Scoring Function}
\label{sec:scoring function}
The purpose of the scoring function is to produce importance scores for each entry in the key $\mathbf{K}$ w.r.t the query $\mathbf{q}$ for the task at hand, which in our case is classification. We experiment with two scoring functions: additive and transformer-based.

\smallskip
\noindent\textbf{Additive: } This is same as the scoring function used in \citet{Bahdanau2015NeuralMT}, where the scores are produced as follows:
\begin{align*}
    s_j = \mathbf{v}^{T} \tanh (\mathbf{W}_q~ \mathbf{q} + \mathbf{W}_k~ \mathbf{k}_j)~, 
\end{align*}

\noindent where, $\mathbf{v} \in \mathbb{R}^m$, $\mathbf{W}_q \in \mathbb{R}^{m\times m}$ and $\mathbf{W}_k \in \mathbb{R}^{m \times n}$ are trainable weights.

\begin{figure}
    \centering
    \includegraphics[height=150pt]{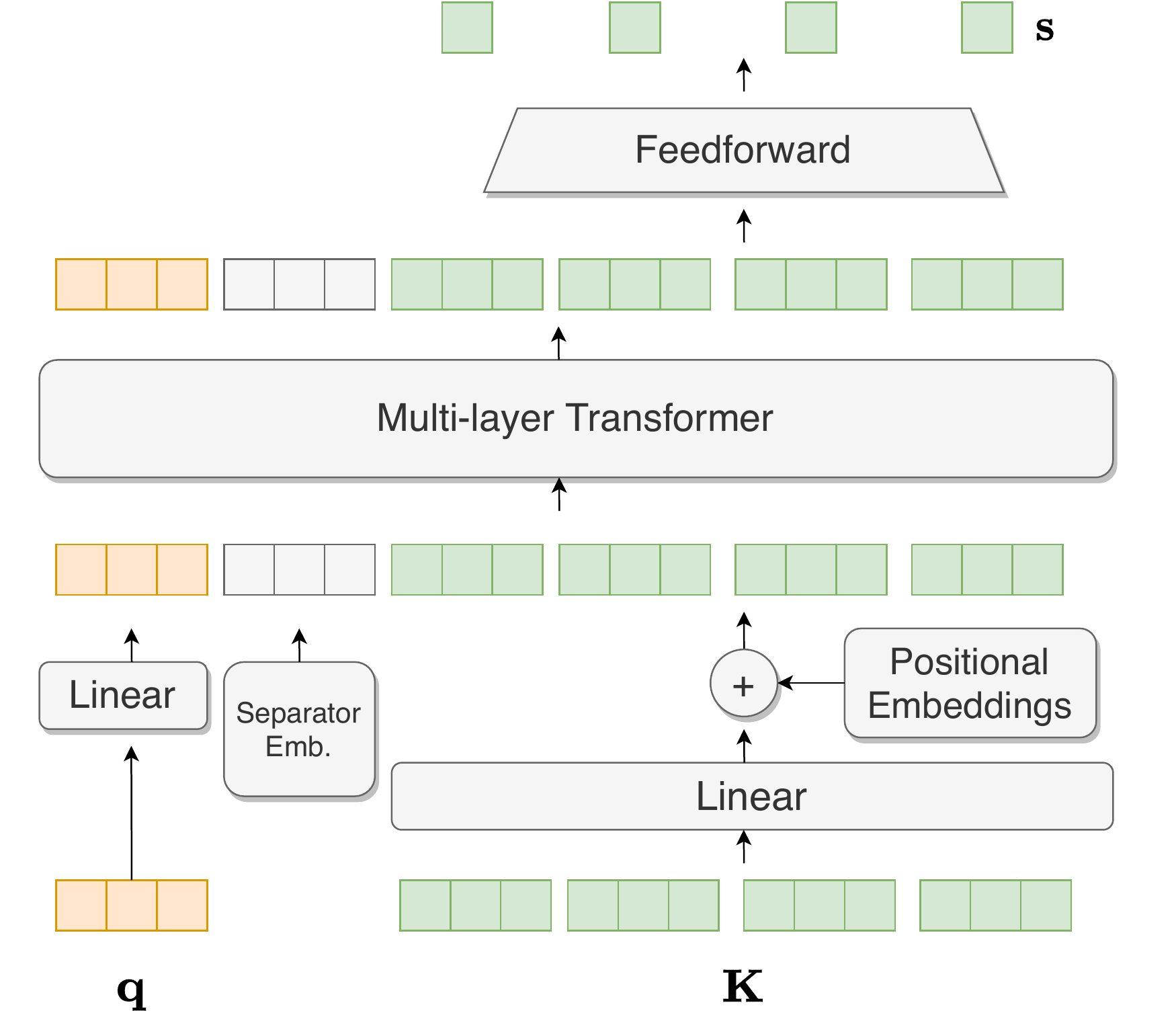}
    \caption{Architecture of TAScore. $\mathbf{q}$ and $\mathbf{K}$ are input query and keys, respectively, and $\mathbf{s}$ are the output scores.}
    \label{fig:transformer scorer}
\end{figure}

\smallskip
\noindent\textbf{Transformer-based Attention Score (TAScore): } While the additive scoring function is simple and easy to train, it suffers from one major drawback in our setting: since we freeze the weights of our embedder and do not use multiple layers of trainable attention (Section \ref{sec:training}), the additive attention can struggle to resolve references -- finding the correct attribute when there are multiple entities of interest, especially when there are multiple distinct medications (Section \ref{sec:effect of scoring}).
% when there are multiple entities of interest (i.e. multiple distinct medications) present in the text. 
For this reason, we propose a novel multi-layer transformer-based attention scoring function (TAScore) which can perform this reference resolution while also preserving the \emph{attention bottleneck}. 
%\sai{comeback again about novelness}% essential for maintaining the faithfulness of attention weights resulting in stable extraction performance. 
 Figure \ref{fig:transformer scorer} shows the architecture of TAScore. The query and key vectors are projected to the same space using two separate linear layers while also adding sinusoidal positional embeddings to the key vectors. A special trainable separator vector is added between the query and key vectors and the entire sequence is passed through a multi-layer transformer \cite{attention-is-all}. Finally, scalar scores (one corresponding to each vector in the key) are produced from the outputs of the transformer by passing them through a feed-forward layer with dropout. 

%\todo[inline]{Add a figure and may be an equation if needed.}

\subsubsection{Projection Function}
\label{sec:projection function}
A projection function $\Pi \colon \mathbb{R}^l \to \Delta^l$ in the context of attention distribution, normalizes the real valued importance scores by projecting them to an $(l-1)$-dimensional probability simplex $\Delta^l$.
\citet{fusedmax} provide a unified view of the projection function as follows:
\vspace{-1.2mm}
\begin{align*}
    \Pi_{\Omega}(\mathbf{s}) = \argmax_{\mathbf{a} \in \Delta^l}~ \mathbf{a}^{T} \mathbf{s} - \gamma \Omega(\mathbf{a})~.
\end{align*}
\noindent Here, $\mathbf{a}\in \Delta^l$, $\gamma$ is a hyperparameter and $\Omega$ is a regularization penalty \newlyadded{which allows us to introduce problem specific inductive bias into our attention distribution}. When $\Omega$ is strongly convex, we have a closed form solution to the projection operation as well as its gradient \cite{fusedmax, learning-with-fenchel}. \newlyadded{Since we use the attention distribution to perform extraction}, we experiment with the following instances of projection functions in this work.

\smallskip
\noindent\textbf{Softmax:}~~ $\Omega(\mathbf{a}) = \sum^l_{i=1} a_i \log a_i$\\
Using the negative entropy as the regularizer, results in the usual softmax projection operator $\Pi_{\Omega}(\mathbf{s}) = \frac{\exp({\mathbf{s}/\gamma})}{\sum^l_{i=1} \exp({s_i/\gamma})}~.$
%\begin{align*}
%    \Pi_{\Omega}(\mathbf{s}) = \frac{e^{\mathbf{s}/\gamma}}{\sum^l_{i=1} e^{s_i/\gamma}}~.
%\end{align*}
%

\smallskip
\noindent\textbf{Fusedmax:}  $\Omega(\mathbf{a}) = \frac{1}{2}\Vert\mathbf{a}\Vert^2_2 + \sum^l_{i=1}\left| a_{i+1} - a_{i}\right|$\\
Using squared loss with fused-lasso penalty \cite{fusedmax}, results in a projection operator which produces sparse as well as contiguous attention weights\footnote{Some example outputs of softmax and fusedmax on random inputs are shown in Appendix \ref{app:examples projection}}. The fusedmax projection operator can be written as $
    \Pi_{\Omega}(\mathbf{s}) = P_{\Delta^l}\left(P_{TV}(s)\right),~~\textrm{where}
$
$$ 
    P_{TV}(\mathbf{s}) = \argmin_{\mathbf{y}\in\mathbb{R}^l}\Vert\mathbf{y}-\mathbf{s}\Vert_2^2 + \sum_{d=1}^{l-1} |y_{d+1} - y_d|
$$
is the proximal operator for 1d Total Variation Denoising problem, and $P_{\Delta^l}$ is the euclidean projection operator. Both these operators can be computed non-iteratively as described in \citet{Condat2013} and \citet{Duchi}, respectively. The gradient of Fusedmax operator can be efficiently computed as described in \citet{fusedmax}.\footnote{The pytorch implementation to compute fusedmax used in this work is available at \url{https://github.com/dhruvdcoder/sparse-structured-attention}.}

\smallskip
\noindent\textbf{Fusedmax*:} We observe that while softmax learns to focus on the right region of text, it tends to assign very low attention weights to some tokens of phrases resulting in multiple discontinuous spans per attribute, while Fusedmax on the other hand, almost always generates contiguous attention weights. However, Fusedmax makes more mistakes in identifying the overall region that contains the target span (Section \ref{sec:effect of projection}). In order to combine the advantages of softmax and Fusedmax, we first train a model using softmax as the projector and then swap the softmax with Fusedmax in the final few epochs. We call this approach Fusedmax*.

\section{Model}
\label{sec:model}
\begin{figure}[]
    \centering
    \includegraphics[height=70mm,width=0.40\textwidth]{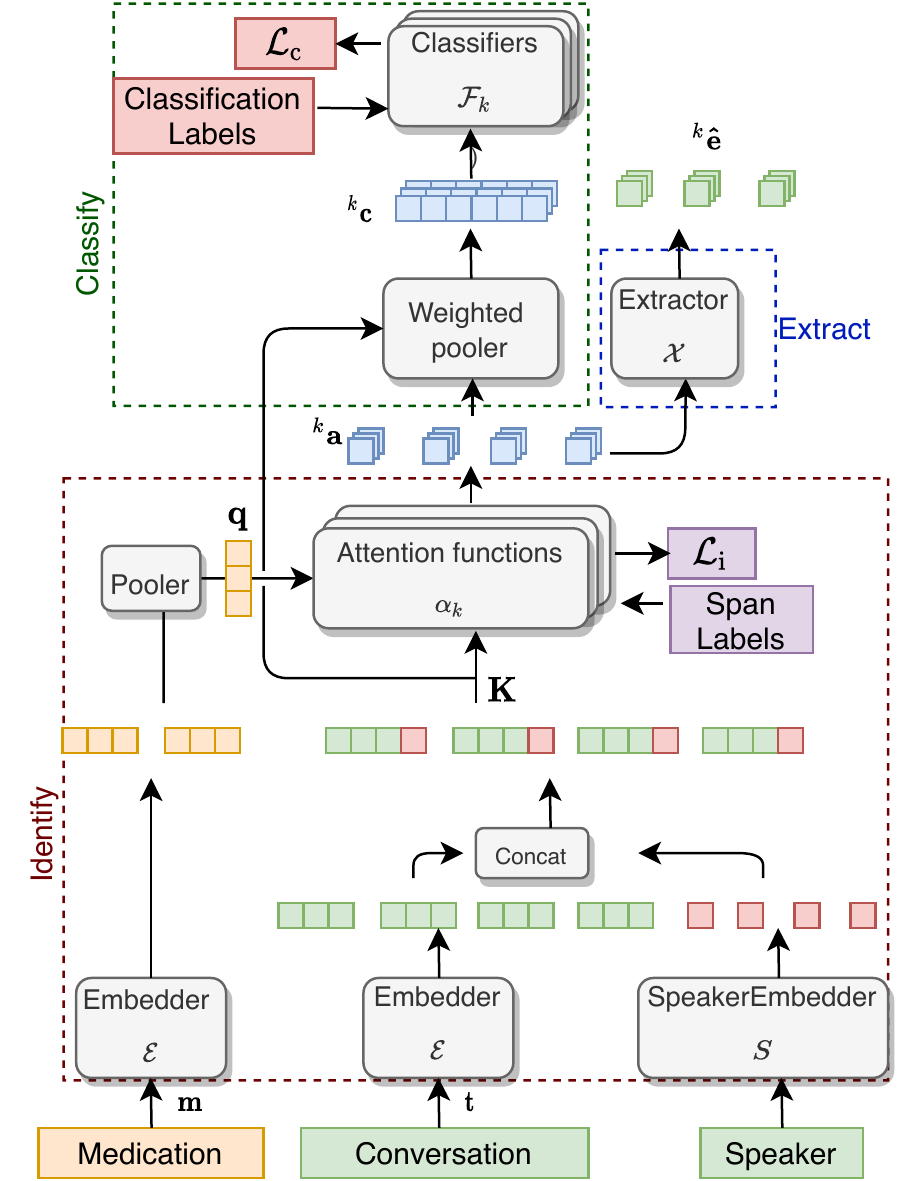}
    \caption{Complete model for weakly supervised MR attribute extraction.}
    \label{fig:model_overview}
\end{figure}

%In the absence of span level labels, we utilize the noisy and normalized classification labels to perform MR attribute extraction. 
Our classification + extraction model uses MR attributes classification labels to extract MR attributes. 
%Our complete model can be divided into two sub-models: classification and extraction, where the classification model is further divided into two phases -- identify and classify.
The model can be divided into three phases: identify, classify and extract (Figure \ref{fig:model_overview}). 
%\sai{Write a line about "extract" part.}
%-- that are separated by an attention bottleneck.
%\todo[inline]{Add a bridge text here to setup the rest of the section.} 
The identify phases encodes the input text and medication name and uses the attention bottleneck to produce attention over the text. Classify phase computes the context vector using the attention from the identify phases and classifies the context vectors. Finally, the extract phase uses the attention from the identify phase to extract spans corresponding to MR attributes. 

% \sai{incorporate this description, use phases}
% %Using this argument as our motivation, we construct a 
% Our \sai{+extraction?}classification model (described in Figure \ref{fig:model_overview}, Section \ref{sec:model}) consisting of contextualized token embedder $\mathcal{E}$, attention bottlenecks $\alpha$, and classifiers $\mathcal{F}$ in that order. $\mathcal{E}$ encodes the input, and 

% Before getting into the details of the model, we introduce
\noindent\textbf{Notation}:
% \smallskip
Let the dataset $\mathcal{D}$ be $\{(\mathbf{x^{(1)}}, \mathbf{y^{(1)}}), \dots (\mathbf{x^{(N)}}, \mathbf{y^{(N)}}) \}$. Each 
$\mathbf{x}$ consists of a medication $m$ and conversation text $\mathbf{t}$, and each $\mathbf{y}$ consists of classification labels for \freq, \route~and \startstop, i.e,  $\mathbf{y} = (^fy, ^ry, ^cy)$, respectively. The number of classes for each attribute is denoted by $^{(\cdot)}n$. As seen from Table \ref{tab:class labels}, $^fn=12$, $^rn=10$ and $^cn=8$.
The length of a text excerpt is denoted by  $l$. The extracted span for attribute $k\in\{f,r,c\}$ is denoted by a binary vector $^{k}\mathbf{e}$ of length $l$, such that $^{k}e_j =1$, if $j^{\text{th}}$ token is in the extracted span for attribute $k$.

\subsection{Identify}
\label{sec:model:identify}
As shown in the Figure \ref{fig:model_overview}, the \emph{identify} phase finds the most relevant parts of the text w.r.t each of the three attributes. For this, we first encode the text as well as the given medication using a contextualized token embedder $\mathcal{E}$. In our case, this is 1024 dimensional BERT \cite{devlin-etal-2019-bert}\footnote{The pre-trained weight for BERT is from the HuggingFace library\cite{huggingface} }. Since BERT uses WordPiece representations \cite{wordpiece}, we average these wordpiece representations to form the word embeddings. In order to supply the speaker information, we concatenate a 2-dimensional fixed vocabulary speaker embedding to every token embedding in the text to obtain speaker-aware word representations.

We then perform average pooling of the medication representations to get a single vector representation for the medication\footnote{Most medication names are single word, however a few medicines have names which are upto 4-5 words.}.  Finally, with the given medication representation as the query and the speaker-aware token representations as the key, we use three separate attention functions (attention bottleneck), one for each attribute (no weight sharing), to produce three sets of normalized attention distributions $^f\mathbf{\hat{a}}$, $^r\mathbf{\hat{a}}$ and $^c\mathbf{\hat{a}}$ over the tokens of the text. The \emph{identify} phase can be succinctly described as follows: 
\begin{align*}
    ^k\mathbf{a} = {^k\alpha}(\mathcal{E}(m), \mathcal{E}(t))~,~~\text{where } k \in \{f, r, c\}
\end{align*}
Here, each $^k\mathbf{\hat{a}}$ is an element of the probability simplex $\Delta^l$ and is used to perform attribute extraction (Section \ref{sec:extract}).

\subsection{Classify}
\label{sec:model:classify}
% Once we have the attention distributions, 
We obtain the attribute-wise context vectors~$^k\mathbf{c}$, as the weighted sum of the encoded tokens ($\mathbf{K}$ in Figure \ref{fig:model_overview}) where the weights are given by the attribute-wise attention distributions $^k\mathbf{a}$.
 %\begin{align*}
 %    ^k\mathbf{c} = \sum^l_{j=1}~{^ka}_j~ \mathbf{K}_j ~,~~\text{where } k \in \{f, r, c\}~.
 %\end{align*}
% Finally, in order t
To perform the classification for each attribute, the attribute-wise context vectors are used as input to feed-forward neural networks $\mathcal{F}_k$ (one per attribute), as shown below:\footnote{Complete set of hyperparameters used is given in Appendix \ref{app:hyperparams}}% $\mathcal{F}\colon \mathbb{R}^n \to \Delta^{^kn}$:
\begin{align*}
    ^k\mathbf{p} &= \text{softmax}\left(\mathcal{F}_k(^k\mathbf{c})\right)\\
    ^k\hat{y} &= \argmax_{j \in \{1,2,\dots, ^kn\}} {^kp_j}~,~~\text{where } k \in \{f, r, c\}.
\end{align*}
\subsection{Extract}
\label{sec:extract}
The spans are extracted from the attention distribution using a fixed extraction function $\mathcal{X}\colon \Delta^l \to \{0,1\}^l$, defined as:
\begin{align*}
 ^k\hat{e}_j=\mathcal{X}_k(^k\mathbf{a})_j &= 
  \begin{cases} 
   1 & \text{if } {^ka}_j > {^k\gamma} \\
   0       & \text{if } {^ka}_j \leq {^k\gamma}~,
  \end{cases}
\end{align*}
%\vspace{-2mm}
\noindent where $^k\gamma$ is the extraction threshold for attribute $k$. 
For softmax projection function, it is important to tune the attribute-wise extraction thresholds $\mathbf{\gamma}$. We tune these using extraction performance on the extraction validation set. For fusedmax projection function which produces spare weights, the thresholds need not be tuned, and hence are set to $0$.

\subsection{Training}
\label{sec:training}
We train the model end-to-end using gradient descent, except the \emph{extract} module (Figure \ref{fig:model_overview}), which does not have any trainable weights, and the embedder $\mathcal{E}$. {Freezing the embedder is vital for the performance, since not doing so results in excessive dispersion of token information to other nearby tokens, resulting in poor extractions.} 

\smallskip
The total loss for the training is divided into two parts as described below.

\noindent\textbf{(1) Classification Loss $\mathcal{L}_c$: } In order to perform classification with highly class imbalanced data (see Table \ref{tab:class labels}), we use weighted cross-entropy: 
\begin{align*}
    \mathcal{L}_c &=   \sum_{k \in \{ f, r, c\}} - ~{^kw_{^ky}}~ \log \left({^kp_{^ky}}\right)~,
\end{align*}
%\vspace{-3mm}
\noindent where the class weights ${^kw_{^ky}}$ are obtained by inverting each class' relative proportion.

\smallskip
\noindent\textbf{(2) Identification Loss $\mathcal{L}_i$: }If span labels $\mathbf{{e}}$ are present for some subset $\mathcal{A}$ of training examples, we first normalize these into ground truth attention probabilities $\mathbf{a}$:
\begin{align*}
    ^ka_j &= \frac{^ke_j}{\sum^l_{j=1} {^ke_j}} ~~~\text{~~for~~} k \in \{f,r,c\}
\end{align*}
\noindent We then use KL-Divergence between the ground truth attention probabilities and the ones generated by the model ($\mathbf{\hat{a}}$) to compute identification loss $\mathcal{L}_i =   \sum_{k \in \{ f, r, c\}} ~ \text{KL}\left(^k\mathbf{{a}} \Big\Vert {^k\mathbf{\hat{a}}}\right)$. Note that $\mathcal{L}_i$ is zero for data-points that do not have span labels.
Using these two loss functions, the overall loss $\mathcal{L} =   \mathcal{L}_c + \lambda \mathcal{L}_i$.

\begin{table*}[]
\centering
\resizebox{\textwidth}{!}{%
\begin{tabular}{@{}lll|l|lll|l|lll|l|llll@{}}
\toprule
\multicolumn{3}{c|}{Model} &
  \multirow{2}{*}{\begin{tabular}[c]{@{}l@{}}Span \\ labels\end{tabular}} &
  \multicolumn{4}{c|}{Token-wise extraction F1} &
  \multicolumn{4}{c|}{LCSF1} &
  \multicolumn{4}{c}{Classification F1} \\ \cmidrule(r){1-3} \cmidrule(l){5-16} 
\multicolumn{1}{c}{Encoder} &
  \multicolumn{1}{c}{Scorer} &
  \multicolumn{1}{c|}{Projector} &
   &
  freq. &
  route &
  change &
  Avg. &
  freq. &
  route &
  change &
  Avg. &
  freq. &
  route &
  \multicolumn{1}{l|}{change} &
  Avg. \\ \midrule
\multicolumn{3}{l|}{\textit{Phrase-based baseline}} &
  - &
  41.03 &
  48.57 &
  10.75 &
  33.45 &
  36.26 &
  50.41 &
  11.54 &
  32.73 &
  -\hide{50.66} &
  -\hide{52.01} &
  \multicolumn{1}{l|}{-\hide{29.35}} &
  -\hide{44.00} \\ \midrule
BERT &
  Additive &
  Softmax &
  0 &
  51.22 &
  46.27 &
  22.81 &
  40.10 &
  39.87 &
  46.40 &
  18.92 &
  35.06 &
  51.51 &
  54.06 &
  \multicolumn{1}{l|}{51.65} &
  52.40 \\
BERT &
  Additive &
  Fusedmax &
  0 &
  47.55 &
  51.31 &
  5.10 &
  34.65 &
  46.39 &
  59.10 &
  4.82 &
  36.77 &
  43.54 &
  42.91 &
  \multicolumn{1}{l|}{9.19} &
  31.88 \\
%BERT &
%  Additive &
%  Fusedmax* &
%  0 &
%  44.24 &
%  49.31 &
%  21.37 &
%  38.31 &
%  42.10 &
%  54.49 &
%  24.63 &
%  40.41 &
%  48.04 &
%  47.81 &
%  \multicolumn{1}{l|}{49.89} &
%  48.58 \\
BERT &
  TAScore &
  Softmax &
  0 &
  66.53 &
  48.96 &
  27.61 &
  47.70 &
  61.49 &
  47.34 &
  22.49 &
  43.77 &
  44.93 &
  51.34 &
  \multicolumn{1}{l|}{46.49} &
  47.58 \\
BERT &
  TAScore &
  Fusedmax &
  0 &
  56.35 &
  44.04 &
  22.07 &
  40.82 &
  61.96 &
  50.27 &
  25.25 &
  45.82 &
  51.95 &
  48.37 &
  \multicolumn{1}{l|}{43.00} &
  47.77 \\\midrule
%BERT &
%  TAScore &
%  Fusedmax* &
%  0 &
%   &
%   &
%   &
%   &
%   &
%   &
%   &
%   &
%   &
%   &
%  \multicolumn{1}{l|}{} &
%   \\ \midrule
BERT &
  Additive &
  Softmax &
  150 &
  61.56 &
  45.08 &
  33.54 &
  46.73 &
  57.90 &
  48.14 &
  28.28 &
  44.77 &
  55.62 &
  52.42 &
  \multicolumn{1}{l|}{50.40} &
  52.81 \\
BERT &
  Additive &
  Fusedmax &
  150 &
  47.05 &
  52.49 &
  27.69 &
  42.41 &
  42.37 &
  57.50 &
  30.63 &
  43.50 &
  54.04 &
  48.40 &
  \multicolumn{1}{l|}{52.28} &
  51.57 \\
BERT &
  Additive &
  Fusedmax* &
  150 &
  65.90 &
  47.30 &
  34.77 &
  49.32 &
  67.15 &
  51.12 &
  36.04 &
  51.30 &
  56.46 &
  42.63 &
  \multicolumn{1}{l|}{50.68} &
  49.93 \\
BERT &
  TAScore &
  Softmax &
  150 &
  66.53 &
  54.35 &
  34.27 &
  {\textbf{51.72}} &
  62.90 &
  53.05 &
  28.33 &
  48.09 &
  50.13 &
  45.86 &
  \multicolumn{1}{l|}{47.16} &
  47.72 \\
BERT &
  TAScore &
  Fusedmax &
  150 &
  58.24 &
  58.09 &
  25.09 &
  47.32 &
  57.93 &
  64.05 &
  26.70 &
  49.56 &
  51.61 &
  53.95 &
  \multicolumn{1}{l|}{43.51} &
  49.69 \\
BERT &
  TAScore &
  Fusedmax* &
  150 &
  66.90 &
  54.85 &
  33.28 &
  51.67 &
  70.10 &
  60.05 &
  35.92 &
  \textbf{55.36} &
  64.26 &
  44.50 &
  \multicolumn{1}{l|}{51.21} &
  \textbf{53.32} \\ \bottomrule
\end{tabular}%
}
\caption{Attribute extraction performance for various combinations of scoring and projection functions. The avg. columns represent the macro average of the corresponding metric across the attributes. }
\label{tab:all results}
\end{table*}

\begin{table*}[]
\centering
\resizebox{\textwidth}{!}{%
\begin{tabular}{@{}l|l|lll|l|llll@{}}
\toprule
\multicolumn{1}{c|}{\multirow{2}{*}{Training Type}} &
  \multicolumn{1}{c|}{\multirow{2}{*}{Model}} &
  \multicolumn{4}{c|}{Tokenwise Extraction F1} &
  \multicolumn{4}{c}{Classification F1} \\ \cmidrule(l){3-10} 
\multicolumn{1}{c|}{}      & \multicolumn{1}{c|}{} & freq. & route & change & avg.  & freq. & route & \multicolumn{1}{l|}{change} & avg.  \\ \midrule
Classification only        & BERT Classifiers      & -     & -     & -      & -     & 74.72 & 40.82 & \multicolumn{1}{l|}{55.76}  & 58.48 \\
Classification only & BERT+TAScore+Fusedmax* &  58.55 & 45.00 & 24.43 & 42.66 & 52.45 & 46.37 & \multicolumn{1}{l|}{43.00} & 47.27 \\
Extraction only            & BERT+TAScore+Fusedmax*  & 53.79 & 44.44 & 14.32  & 37.18 & -     & -     & \multicolumn{1}{l|}{-}      & -     \\
Classification +Extraction & BERT+TAScore+Fusedmax*  & 66.90 & 54.85 & 33.28  & 51.67 & 64.26 & 44.50 & \multicolumn{1}{l|}{51.21}  & 53.32 \\ \bottomrule
\end{tabular}%
}
\caption{Effect of performing extraction+classification jointly in our proposed model. While the \textit{Extraction Only} training only uses the 150 examples which are explicitly annotated with span labels, the \textit{Classification only} training uses the complete training dataset with classification labels.}
\label{tab:effect joint model}
\end{table*}

\section{Metrics}
\label{sec:metrics}

\noindent\textbf{Token-wise F1 (TF1)}: 
Each token in text is either part of the extracted span (positive class) for an attribute or not (negative class). 
Token-wise F1 score is the F1 score of the positive class obtained by considering all the tokens in the dataset as separate binary classification data points. TF1 is calculated separately for each attribute.

\smallskip
\noindent\textbf{Longest Common Substring F1 (LCSF1):} LCSF1 measures if the extracted spans, along with being part of the gold spans, are contiguous or not. Longest Common Substring (LCS) is the longest overlapping contiguous span of tokens between the predicted and gold spans. LCSF1 is defined as the harmonic mean of LCS-Recall and LCS-Precision which are defined per extraction as:

\vspace{-3mm}
{\small
\begin{align*}
    \text{LCS-Recall} &= \frac{\text{\#tokens in LCS}}{\text{\#tokens in gold span}}\\
    \text{LCS-Precision} &= \frac{\text{\#tokens in LCS}}{\text{\#tokens in predicted span}}
\end{align*}
}%
\section{Results and Analysis}
\label{sec: results and analysis}
% \subsection{Results}
\label{sec:results}

Table \ref{tab:all results} shows the results obtained by various combinations of attention scoring and projection functions on the task of MR attribute extraction in terms of the metrics defined in Section \ref{sec:metrics}. 
It also shows the classification F1 score to emphasize how the attention bottleneck affects classification performance.
The first row shows how a simple phrase based extraction system would perform on the task.\footnote{The details about the phrase based baseline are presented in Appendix \ref{app:phrase based}}

\subsection{Effect of Span labels}

In order to see if having a small number of extraction training data-points (containing explicit span labels) helps the extraction performance, we annotate 150 (see Section \ref{sec:task and data} for how we sampled the datapoints) of the training data-points with span labels. As seen from Table \ref{tab:all results}, even a small number of examples with span labels ($\approx 0.3 \%$) help a lot with the extraction performance for all models. We think this trend might continue if we add more training span labels. We leave the finding of the right balance between annotation effort and extraction performance as a future direction to explore.

\subsection{Effect of classification labels}

In order to quantify the effect of performing the auxiliary task of classification along with the main task of extraction, we train the proposed model in three different settings. (1) The \textit{Classification Only} uses the complete dataset (\texttildelow45k) but only with the classification labels. (2) The \textit{Extraction Only} setting only uses the 150 training examples that have span labels. (3) Finally, the \textit{Classification+Extraction} setting uses the 45k examples with classification labels along with the 150 examples with the span labels to train the model.
Table \ref{tab:effect joint model} (rows 2, 3 and 4) shows the effect of having classification labels and performing extraction and classification jointly using the proposed model. The model structure and the volume of the classification data (\texttildelow45k examples) 
%compared to the extraction data (150 examples) 
makes the auxiliary task of classification extremely helpful for the main task of extraction, even with the presence of label noise.

It is worth noting that the classification performance of the proposed method is also improved by explicit supervision to the extraction portion of the model (row 2 vs 4, Table \ref{tab:effect joint model}).
In order to set a reference for classification performance, we train strong classification only models, one for each attribute, using pretrained BERT.
These \textit{BERT Classifiers}, are implemented as described in \citet{devlin-etal-2019-bert} with input consisting of the text and medication name separated by a [SEP] token (row 1). Based on the improvements achieved in the classification performance using span annotations, we believe that having more span labels can further close the gap between the classification performance of the proposed model and the BERT Classifiers. However, this work focuses on extraction performance, hence improving the classification performance is left to future work.

\subsection{Effect of projection function}
\label{sec:effect of projection}
\begin{figure}[h!]
\centering
\subfloat[BERT+TAScore+Fusedmax*]{\includegraphics[width = \columnwidth]{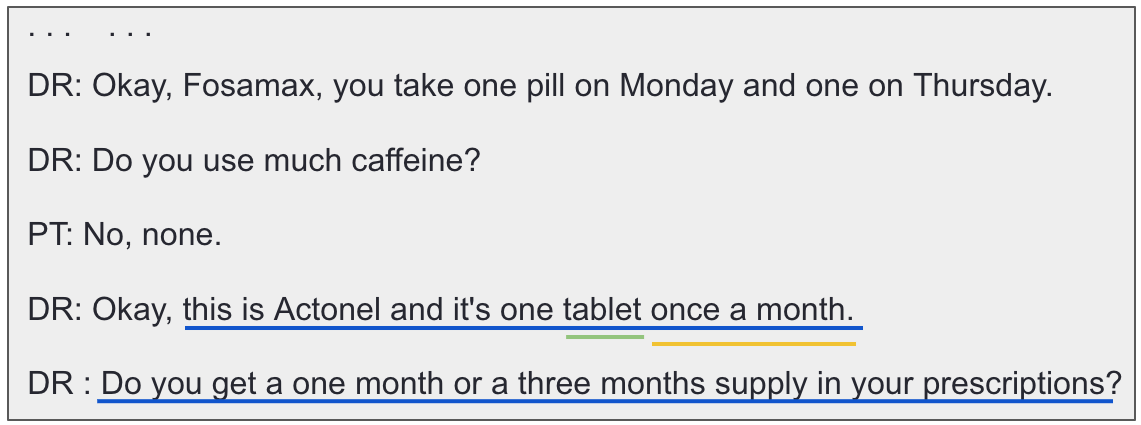}}\\[-1mm]
\subfloat[BERT+TAScore+Softmax]{\includegraphics[width = \columnwidth]{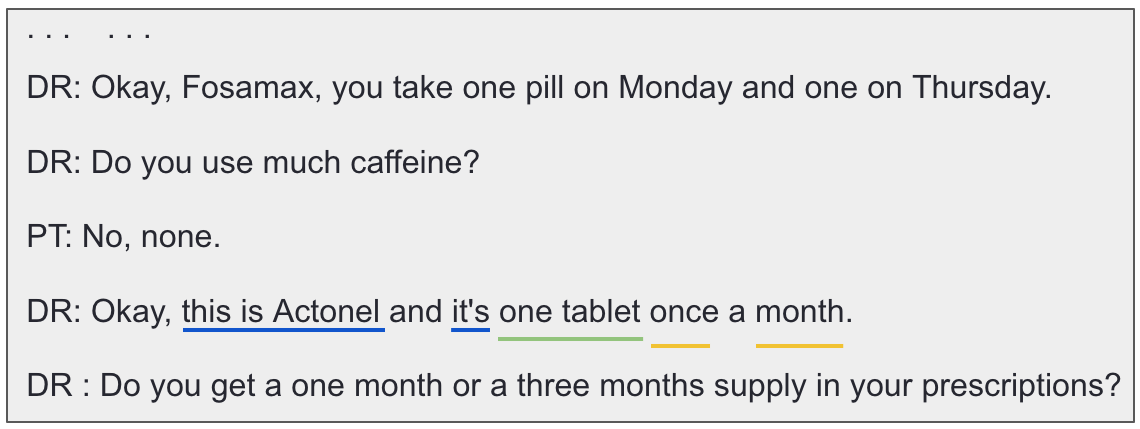}}
\vspace{-1mm}
\caption{Difference in extracted spans for MR attributes with models that uses Fusedmax* and Softmax, for the medication Actonel. Blue: \startstop, green: \route~and yellow: \freq. Refer Figure \ref{fig:dialogue} for ground-truth annotations.}
\label{fig:extraction examples}
\end{figure}

While softmax with post-hoc threshold tuning achieves consistently higher TF1 compared to fusedmax (which does not require threshold tuning), the later achieves better LCSF1.  
We observe that while the attention function using softmax projection focuses on the correct portion of the text, it drops intermediate words, resulting in multiple discontinuous spans. Fusedmax on the other hand almost always produces contiguous spans.
Figure \ref{fig:extraction examples} further illustrates this point using a test example.
%However, models with fusedmax have difficulties in training\TODO{sai: be more specific, did it need hyperparameter tunning or something? or say 'in general' and provide citation}.
%In order to combine the advantages of softmax and fusedmax, we first train a model using softmax as the projector and then swap softmax with fusedmax in the last few epochs resulting in the model which performs well on both metrics TF1 as well as LCSF1. This model is denoted using Fusedmax* in Table \ref{tab:all results}.
The training trick which we call fusedmax* swaps the softmax projection function with fusedmax during the final few epochs to combine the strengths of both softmax and fusedmax. This achieves high LCSF1 as well as TF1.

%\begin{figure}[]
%    \centering
%    \includegraphics[width=\columnwidth]{images/bert-transformer-fusedmax-from-softmax.png}
%    \caption{Extracted spans for MR attributes using BERT+Transformer+Fusedmax* model. Here the given medication is Actonel. Blue is the span for \startstop, green for \route~and yellow for \freq.}
%    \label{fig:extraction fusedmax}
%\end{figure}
%
%\begin{figure}
%    \centering
%    \includegraphics[width=\columnwidth]{images/bert-transformer-softmax.png}
%    \caption{Extracted spans for MR attributes using BERT+Transformer+Softmax model. Here the given medication is Actonel. Blue is the span for \startstop, green for \route~and yellow for \freq. {Sai: this looks good need some large span examples for freq and route as well(just of fusedmax)}}
%    \label{fig:extraction softmax}
%\end{figure}

\subsection{Effect of scoring function}
\label{sec:effect of scoring}
Table \ref{tab:effect of scorer} shows the percent change in the extraction F1 if we use TAScore instead of additive scoring (everything else being the same). As seen, there is a significant improvement irrespective of the projection function being used. 
\begin{table}[h!]
\centering
\resizebox{\columnwidth}{!}{%
\begin{tabular}{@{}l|lll|lll@{}}
\toprule
\multirow{2}{*}{\begin{tabular}[c]{@{}l@{}}Scorer\end{tabular}} & \multicolumn{3}{l|}{TF1 ($\Delta$\%)} & \multicolumn{3}{l}{LCSF1 ($\Delta$\%)} \\ \cmidrule(l){2-7} 
 &
  \begin{tabular}[c]{@{}l@{}}MM\\ (77.3)\end{tabular} &
  \begin{tabular}[c]{@{}l@{}}SM\\ (22.7)\end{tabular} &
  \begin{tabular}[c]{@{}l@{}}All\\ (100)\end{tabular} &
  \begin{tabular}[c]{@{}l@{}}MM\\ (77.3)\end{tabular} &
  \begin{tabular}[c]{@{}l@{}}SM\\ (22.7)\end{tabular} &
  \begin{tabular}[c]{@{}l@{}}All\\ (100)\end{tabular} \\ \midrule
softmax                                                                        & +11.1        & +10.6       & +10.6       & +6.5         & +6.6         & +6.3        \\
fusedmax                                                                       & +12.1        & +8.3        & +11.5       & +16.4        & +15.5        & +13.9       \\
fusedmax*                                                                      & +5.4         & +1.9        & +4.7        & +9.25        & +1.1         & +7.9        \\ \bottomrule
\end{tabular}%
}
\caption{MR extraction improvement (\%) brought by TAScore over additive scorer in the full test set (All=100\%), and test subset with single medication (SM=22.7\%) and multiple medications (MM=77.3\%) in the \emph{text}.}
\label{tab:effect of scorer}
\end{table}

The need for TAScore stems from the difficulty of the additive scoring function to resolve references between spans when there are multiple medications present. In order to measure the efficacy of TAScore for this problem, we
% perform string matching for each token in each data-point against the set of all medications in the dataset, and\sai{Remove?:} 
divide the test set into two subsets: data-points which have multiple distinct medications in their text (MM) and data-points that have single medication only. 
As seen from the first two columns for both the metrics in Table \ref{tab:effect of scorer}, using TAScore instead of additive results in more improvement in the MM-subset compared to the SM-subset, showing that using transformer scorer does help with resolving references when multiple medications are present in the text.

\begin{figure}
    \centering
    \includegraphics[width=.9\columnwidth]{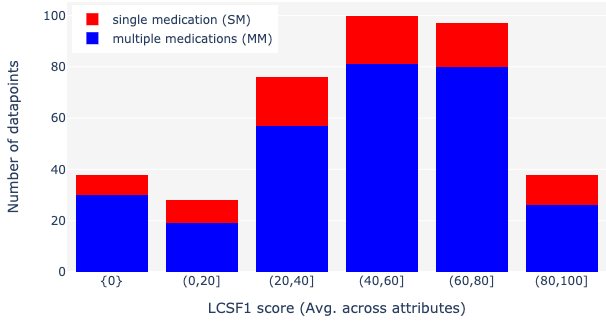}
    \caption{Distribution of the Avg. LCSF1 for the best performing model (BERT+TAScore+Fusedmax*). A significant number ($\approx10\%$) of datapoints with multiple medication in their \emph{text} get LCSF1 of zero (1st bar).}
    \label{fig:LCSF1 distribution}
\end{figure}

Figure \ref{fig:LCSF1 distribution} shows the distribution of Avg. LCSF1 (average across all three attributes). It can be seen that there are a significant number of datapoints in the MM subset which get LCSF1 of zero, showing that even when the transformer scorer achieves improvement on MM subset, it gets quite a lot of these data-points completely wrong. This shows that the there is still room for improvement.

\subsection{Discussion}
In summary, our analysis reveals that Fusedmax/Fusedmax* favors contiguous extraction spans which is a necessity for our task. 
Irrespective of the projection function used, the proposed scoring function TAScore improves the extraction performance when compared to the popular additive scoring function.
%The combination of TAScore and Fusedmax* gets non-zero LCSF1 for 90\% of the test set, i.e, 90\% of the time the predicted extractions have some overlap with the annotated extractions.
The proposed model architecture is able to establish a synergy between the classification and span extraction tasks where one improves the performance of the other.
Overall, the proposed combination of TAScore and Fusedmax* achieves a 22 LCSF1 points improvement over the phrase-based baseline and 10 LCSF1 points improvement over the naive additive and softmax combination.

%\begin{figure}[]
%\begin{tcolorbox}[width=0.45\textwidth]
%\begin{doublespacing}
%    \begin{dialogue}
%    \small
%        \speak{DR}  Limiting your alcohol consumption is important, so, and, um, so, you know, I would recommend \hmed{vitamin D\textsuperscript{1}} \hstartstop{to be taken\textsuperscript{1}}. Have you had \hmed{Fosamax\textsuperscript{2}} before?
%        \speak{PT}  I think my mum did.
%        \speak{DR} Okay, \hmed{Fosamax\textsuperscript{2}}, \hstartstop{you take\textsuperscript{2}} \hfreq{one \hroute{pill\textsuperscript{2}} on Monday and one on Thursday\textsuperscript{2}}.
%        \speak{DR} Do you use much caffine?
%        \speak{PT} No, none.
%        \speak{DR} Okay, \hstartstop{this is\textsuperscript{3}} \hmed{Actonel\textsuperscript{3}} and it's \hfreq{one \hroute{tablet\textsuperscript{3}} once a month\textsuperscript{3}}.
%        \speak{DR} Do you get a one month or a three months supply in your prescriptions?
%    \end{dialogue}
%\end{doublespacing}
%\end{tcolorbox}
%\caption{An example excerpt from a doctor-patient conversation transcript. Here, there are three \hmed{medications} mentioned indicated by the superscript. The extracted attributes, \hstartstop{change}, \hroute{route} and \hfreq{frequency}, for each medications are also shown. }
%\label{fig:dialogue}
%\end{figure}
\section{Related Work}
\label{sec:related work}

%\points{
%\begin{enumerate}
%    \item weak supervision
%    \item explainablility/rational literature, 1. attention, fusedmax, other similar works in nlp
%    \item related medical conversation papers \cite{du2019learning} \cite{selvaraj2019medication}
%    \item similar task in other areas?
%\end{enumerate}
%}

Existing literature directly related to our work can be bucketed into two categories -- related methods and related tasks.

\smallskip
\noindent\textbf{Methods: }The recent work on generating rationales/explanations for deep neural network based classification models \cite{Lei2016, Bastings2020, Paranjape2020} is closely related to ours in terms of the methods used. Most of these works use binary latent variables to perform extraction as an intermediate step before classification.
%and employ efficient inference techniques like REINFORCE  \cite{reinforce} or reparameterization trick \cite{gumblesoftmax,kingma2013auto} to perform inference for extraction.
Our work is closely related to \cite{Jain2020,zhong2019finegrained}, who use attention scores to generate rationales for classification models.
%They first obtain saliency scores from a complex classification model like BERT and then use fixed heuristic based extractor (ex: top-k contiguous tokens) to generate rationales. 
These works, however, focus on generating \emph{faithful} and \emph{plausible} explanation for classification as opposed to extracting the spans for attributes of an entity, which is the focus of our work. Moreover, our method can be generalized to any number of attributes while all these methods would require a separate model for each attribute.

\smallskip
\noindent\textbf{Tasks: }Understanding doctor-patient conversations is starting to receive attention recently \cite{rajkomar2019automatically, ben2020soap}. 
%Two works in the area explore MR extraction task. 
%First, 
\citet{selvaraj2019medication} performs MR extraction by framing the problem as a generative question answering task. %While that might be a reasonable approach, 
This approach is not efficient at inference time -- it requires one forward pass for each attribute. 
Moreover, unlike a span extraction model, the generative model might produce hallucinated facts. \citet{du2019learning} obtain MR attributes as spans in text; however, they use a fully supervised approach which requires a large dataset with span-level labels.

 \section{Conclusion and Future work}
\label{sec:conclusion}

We provide a framework to perform MR attribute extraction from medical conversations with weak supervision using noisy classification labels. 
This is done by creating an attention bottleneck in the classification model and performing extraction using the attention weights. 
After experimenting with several variants of attention scoring and projection functions, we show that the combination of our transformer-based attention scoring function (TAScore) combined with Fusedmax* achieves significantly higher extraction performance compared to the other attention variants and a phrase-based baseline.

While our proposed method achieves good performance, there is still room for improvement, especially for text with multiple medications. Data augmentation by swapping or masking medication names is worth exploring. 
An alternate direction of future work involves improving the naturalness of extracted spans. Auxiliary supervision using a language modeling objective would be a promising approach for this. 
%We also hope to apply the proposed approaches on other medical information extraction tasks.

\section*{Acknowledgments}
We thank University of Pittsburgh Medical Center (UPMC) and Abridge AI Inc. for providing access to the de-identified data corpus. 

\bibliography{anthology,emnlp2020}
\bibliographystyle{acl_natbib}

\appendix

\section{Appendices}
\label{sec:appendix}

\subsection{Data}
\label{app:data}

The complete set of \emph{normalized} classification labels for all three medication attributes and their meaning is shown in Table \ref{tab:class labels detailed}. 

Average statistics about the dataset are shown in Table \ref{tab:dataset description stats}.

\begin{table}[!ht]
\centering
%\resizebox{\textwidth}{!}{%
\begin{tabular}{@{}lllll@{}}
\toprule
                   & min & max & mean & $\sigma$ \\ \midrule
\#utterances in \emph{text}    & 3   & 20  & 7.8  & 2.3      \\
\#words in \emph{text}       & 12  & 565 & 80.8 & 41.0     \\
\#words in \emph{freq} span   & 1   & 21  & 4.4  & 2.6      \\
\#words in \emph{route} span  & 1   & 9   & 1.5  & 1.0      \\
\#words in \emph{change} span & 1   & 34  & 6.8  & 4.9      \\ \bottomrule
\end{tabular}%
%}
\caption{Statistics of extraction labels (\#words) and the corresponding \emph{text} }
\label{tab:dataset description stats}
\end{table}

\begin{table*}[!ht]
\centering
\resizebox{\textwidth}{!}{%
\begin{tabular}{@{}|l|l|l|l|@{}}
\toprule
Attribute                 & class  &   Meaning &   class proportion                                                                                        \\ \midrule
\freq &
  \begin{tabular}[c]{@{}l@{}}
  Daily\\ Every morning\\ At Bedtime\\ Twice a day\\ Three times a day\\ Every six hours\\ Every week\\ Twice a week\\ Three times a week\\ Every month\\ Other\\ None\end{tabular} &
   \begin{tabular}[c]{@{}l@{}}
  Take the medication once a day (specific time not mentioned).\\ Take the medication once every morning.\\ At Bedtime\\ Twice a day\\ Three times a day\\ Every six hours\\ Every week\\ Twice a week\\ Three times a week\\ Every month\\ Other\\ None\end{tabular} &
  \begin{tabular}[c]{@{}l@{}}
  8.0 \\  0.9\\  1.7\\  6.5\\  1.6\\  0.2\\  0.9\\  0.2\\  0.3\\  0.3\\  1.5\\  77.9\end{tabular} 
  \\\midrule
\route &
  \begin{tabular}[c]{@{}l@{}}Pill\\ Injection\\ Topical cream\\ Nasal spray\\ Medicated patch\\ Ophthalmic solution\\ Inhaler\\ Oral solution\\ Other\\ None\end{tabular} &
 \begin{tabular}[c]{@{}l@{}}Pill\\ Injection\\ Topical cream\\ Nasal spray\\ Medicated patch\\ Ophthalmic solution\\ Inhaler\\ Oral solution\\ Other\\ None\end{tabular} &
  \begin{tabular}[c]{@{}l@{}}6.8\\  3.5\\  1.0\\  0.5\\  0.2\\  0.2\\  0.2\\   0.1\\  2.1\\  85.5\end{tabular} 
  \\\midrule
\startstop & \begin{tabular}[c]{@{}l@{}}Take\\ Stop\\ Increase\\ Decrease\\ None\\ Other\end{tabular}&
\begin{tabular}[c]{@{}l@{}}Take\\ Stop\\  Increase\\ Decrease\\ None\\ Other\end{tabular} &
\begin{tabular}[c]{@{}l@{}}83.1\\  6.5\\  5.2\\   2.0\\  1.6\\  1.4\end{tabular}
\\ \bottomrule
\end{tabular}
}
\caption{Complete set of normalized classification labels for all three medication attributes and their explanation}
\label{tab:class labels detailed}
\end{table*}

\subsection{Hyperparameters}
\label{app:hyperparams}
We use AllenNLP \cite{allennlp} to implement our models and Weights\&Biases \cite{wandb} to manage our experiments.
Following is the list of hyperparameters used in our experiments:
\begin{enumerate}
    \item \textbf{Contextualized Token Embedder: } We use 1024-dimensional 24-layer \texttt{bert-large-cased} obtained as a pre-trained model from HuggingFace\footnote{\url{https://huggingface.co/bert-large-cased}}. We freeze the weights of the embedder in our training. The max sequence length is set to 256.
    
    \item \textbf{Speaker embedding: } 2-dimensional trainable embedding with vocabulary size of 4 as we only have 4 unique speakers in our dataset: doctor, patient, caregiver and nurse.
    
    \item \textbf{Softmax and Fusedmax: }The temperatures of softmax and fusedmax are set to a default value of 1. The sparsity weight of fusedmax is also set to its default value of 1 for all attributes.
    
    \item \textbf{TAScore: } The transformer used in TAScore is a 2-layer transformer encoder where each layer is implemented as in \citet{attention-is-all}. Both the hidden dimensions inside the transformer (self-attention and feedforward) are set to 32 and all the dropout probabilities are set to 0.2. The linear layer for the query has input and output dimensions of 1024 and 32, respectively. Due to the concatenation of speaker embedding, the linear layer for keys has input and output dimensions of 1026 and 32, respectively. The feedforward layer (which generates scalar scores for each token) on top of the transformer is 2-layered with relu activations and hidden sizes (16, 1).
    
    \item \textbf{Classifiers: } The final classifier for each attribute is a 2-layer feedforward network with hidden sizes  (512, ``number of classes for the attribute'') and dropout probability of 0.2.
\end{enumerate}

\subsection{Examples: Projection Functions}
\label{app:examples projection}
\begin{figure*}
\centering
\subfloat[Positive and negative scores]{\includegraphics[width = \textwidth]{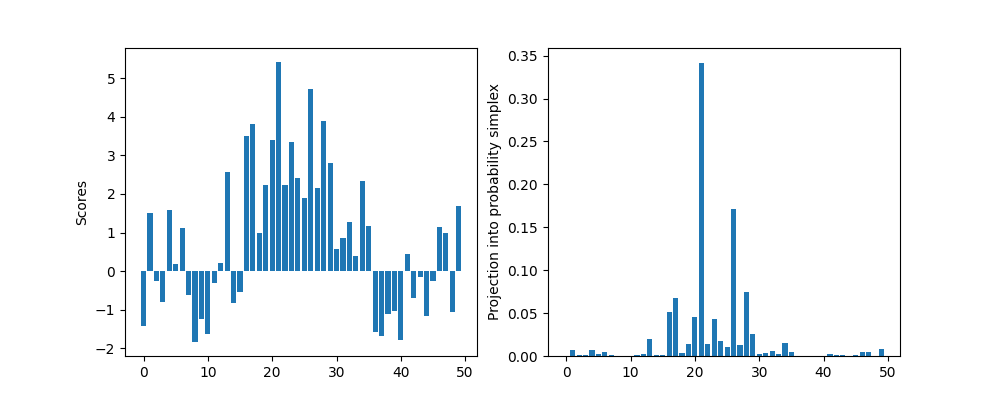}}\\
\subfloat[Positive scores only]{\includegraphics[width = \textwidth]{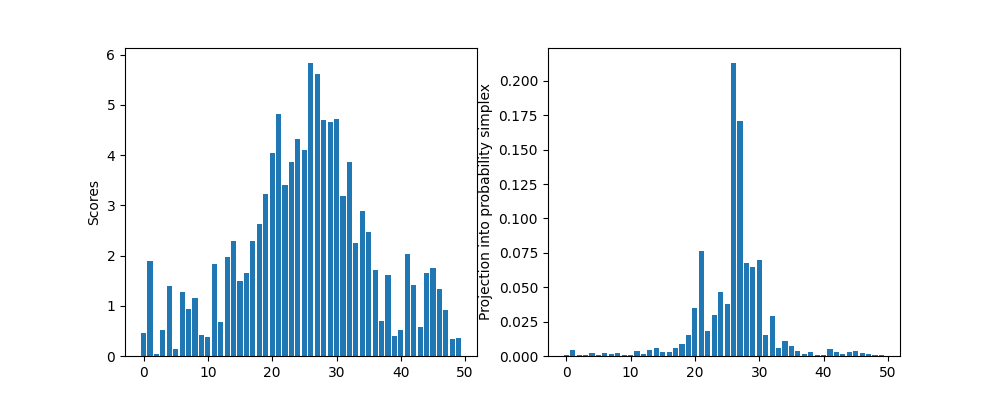}}\\
\subfloat[More uniformly distributed positive scores]{\includegraphics[width = \textwidth]{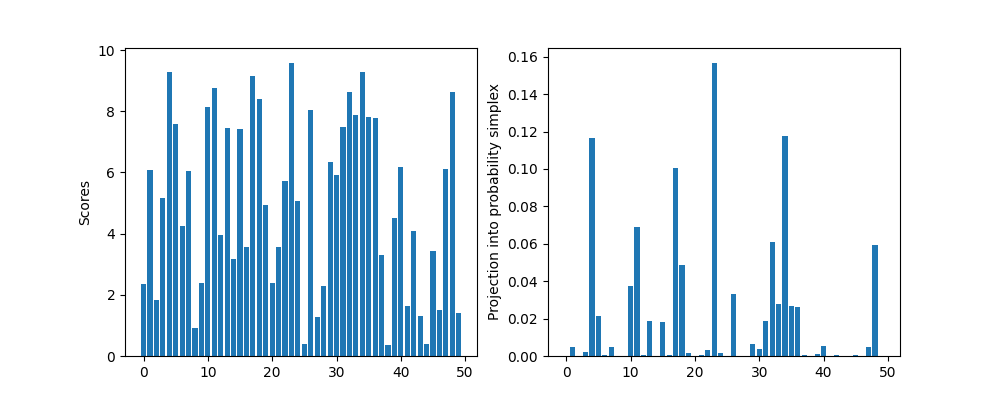}}
\caption{Sample outputs (right column) of softmax function on random input scores (left column).}
\label{fig:examples softmax}
\end{figure*}

\begin{figure*}
\centering
\subfloat[Positive and negative scores]{\includegraphics[width = \textwidth]{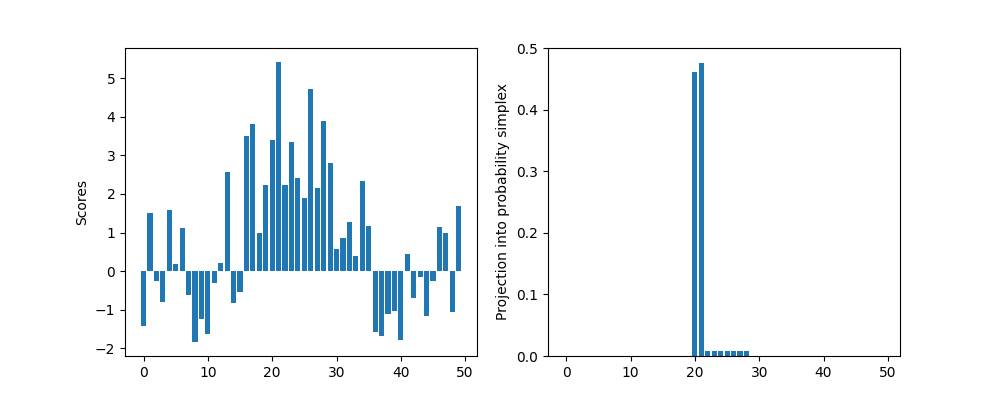}}\\
\subfloat[Positive scores only]{\includegraphics[width = \textwidth]{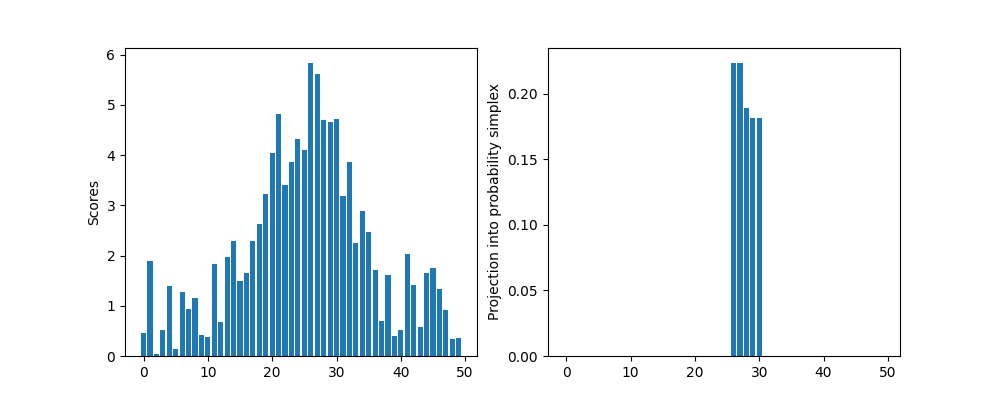}}\\
\subfloat[More uniformly distributed positive scores]{\includegraphics[width = \textwidth]{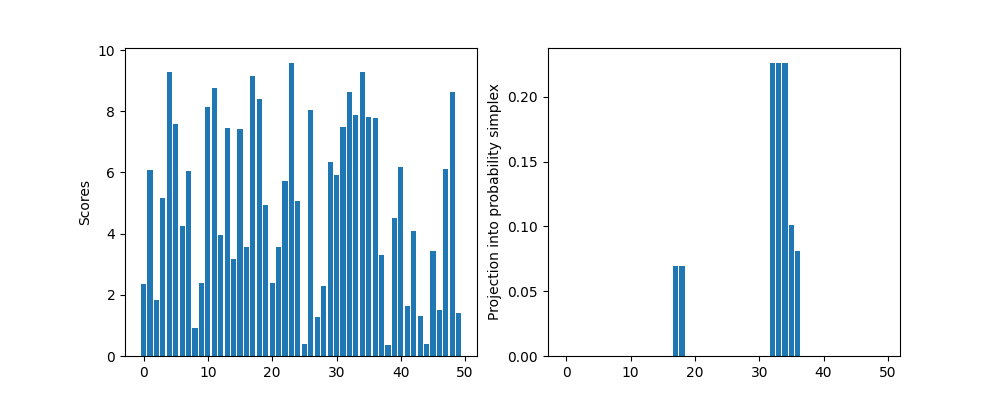}}
\caption{Sample outputs (right column) of fusedmax function on random input scores (left column).}
\label{fig:examples fusedmax}
\end{figure*}

Figures \ref{fig:examples softmax} and \ref{fig:examples fusedmax} show examples of outputs of projection functions softmax and fusedmax on random input scores. 

% \begin{figure}[]
% \begin{tcolorbox}[width=0.45\textwidth]
% \begin{singlespacing}
%     \begin{dialogue}
%     \tiny
%         \speak{DR}  Limiting your alcohol consumption is important, so, and, um, so, you know, I would recommend \hmed{vitamin D\textsuperscript{1}} \hstartstop{to be taken\textsuperscript{1}}. Have you had \hmed{Fosamax\textsuperscript{2}} before?
%         \speak{PT}  I think my mum did.
%         \speak{DR} Okay, \hmed{Fosamax\textsuperscript{2}}, \hstartstop{you take\textsuperscript{2}} \hfreq{one \hroute{pill\textsuperscript{2}} on Monday and one on Thursday\textsuperscript{2}}.
%         \speak{DR} Do you use much caffine?
%         \speak{PT} No, none.
%         \speak{DR} Okay, \hstartstop{this is\textsuperscript{3}} \hmed{Actonel\textsuperscript{3}} and it's \hfreq{one \hroute{tablet\textsuperscript{3}} once a month\textsuperscript{3}}.
%         \speak{DR} Do you get a one month or a three months supply in your prescriptions?
%     \end{dialogue}
% \end{singlespacing}
% \end{tcolorbox}
% \caption{An example excerpt from a doctor-patient conversation transcript. Here, there are three \hmed{medications} mentioned indicated by the superscript. The extracted attributes, \hstartstop{change}, \hroute{route} and \hfreq{frequency}, for each medications are also shown. }
% \label{fig:dialoguerough}
% \end{figure}

\subsection{Phrase based extraction baseline}
\label{app:phrase based}

We implement a phrase based extraction system to provide a baseline for the extraction task.
A lexicon of relevant phrases is created for each class for each attribute as shown in Table \ref{tab:phrase based phrases}. 
We then look for string matches within these phrases and the text for the data-point. If there are matches then the longest match is considered as an extraction span for that attribute.

\begin{table*}[]
\centering
%\resizebox{\textwidth}{!}{%
\begin{tabular}{|l|l|l|} 
\hline
Attribute               & Class               & Phrases                                                                                                                                                                                                \\ 
\hline
\multirow{11}{*}{freq}  & Every Morning       & everyday in the morning \textbar{} every morning \textbar{} morning~                                                                                                                                   \\ 
\cline{2-3}
                        & At Bedtime          & \begin{tabular}[c]{@{}l@{}}everyday before sleeping \textbar{} everyday after dinner \textbar{}\\~every night \textbar{} after dinner \textbar{}\\~at bedtime \textbar{} before sleeping\end{tabular}  \\ 
\cline{2-3}
                        & Twice a day         & \begin{tabular}[c]{@{}l@{}}twice a day \textbar{} 2 times a day \textbar{} two times a day \textbar{}\\~2 times per day \textbar{} two times per day\end{tabular}                                      \\ 
\cline{2-3}
                        & Three times a day   & 3 times a day \textbar{} 3 times per day \textbar{} 3 times every day                                                                                                                                  \\ 
\cline{2-3}
                        & Every six hours     & every 6 hours \textbar{} every six hours                                                                                                                                                               \\ 
\cline{2-3}
                        & Every week          & every week \textbar{} weekly \textbar{} once a week                                                                                                                                                    \\ 
\cline{2-3}
                        & Twice a week        & \begin{tabular}[c]{@{}l@{}}twice a week \textbar{} two times a week \textbar{}~\\2 times a week \textbar{} twice per week \textbar{} two times per week \textbar{}~\\2 times per week\end{tabular}     \\ 
\cline{2-3}
                        & Three times a week  & 3 times a week \textbar{} 3 times per week                                                                                                                                                             \\ 
\cline{2-3}
                        & Every month         & every month \textbar{} monthly \textbar{} once a month                                                                                                                                                 \\ 
\cline{2-3}
                        & Other               &                                                                                                                                                                                                        \\ 
\cline{2-3}
                        & None                &                                                                                                                                                                                                        \\ 
\hline
\multirow{9}{*}{route}  & Pill                & tablet \textbar{} pill \textbar{} capsule \textbar{} mg                                                                                                                                                \\ 
\cline{2-3}
                        & Injection           & pen \textbar{} shot \textbar{} injector \textbar{} injection \textbar{} inject                                                                                                                         \\ 
\cline{2-3}
                        & Topical cream       & cream \textbar{} gel \textbar{} ointment \textbar{} lotion                                                                                                                                             \\ 
\cline{2-3}
                        & Nasal spray         & spray \textbar{} nasal                                                  conversation  transcript.                                                                                                                                 \\ 
\cline{2-3}
                        & Medicated patch     & patch                                                                                                                                                                                                  \\ 
\cline{2-3}
                        & Ophthalmic solution & ophthalmic \textbar{} drops \textbar{} drop                                                                                                                                                            \\ 
\cline{2-3}
                        & Oral solution       & oral solution                                                                                                                                                                                          \\ 
\cline{2-3}
                        & Other               &                                                                                                                                                                                                        \\ 
\cline{2-3}
                        & None                &                                                                                                                                                                                                        \\ 
\hline
\multirow{6}{*}{change} & Take                & take \textbar{} start \textbar{} put you on \textbar{} continue                                                                                                                                        \\ 
\cline{2-3}
                        & Stop                & stop \textbar{} off                                                                                                                                                                                    \\ 
\cline{2-3}
                        & Increase            & increase                                                                                                                                                                                               \\ 
\cline{2-3}
                        & Decrease            & reduce \textbar{} decrease                                                                                                                                                                             \\ 
\cline{2-3}
                        & Other               &                                                                                                                                                                                                        \\ 
\cline{2-3}
                        & None                &                                                                                                                                                                                                        \\
\hline
\end{tabular}
%}
\caption{Phrases used in the phrase based baseline. These are also the most frequently occurring phrases in the free-form annotations.}
\label{tab:phrase based phrases}
\end{table*}

\end{document}